# Optik Karakter Tanıma Sonrası Metin İşleme ile Özgeçmişlerden Bilgi Çıkarımı
# Resume Information Extraction via Post-OCR Text Processing

Selahattin Serdar Helli, Senem Tanberk, Sena Nur Cavsak
serdar.helli1@huawei.com, 0000-0003-1668-0365, sena.nur.cavsak@huawei.com
Research and Innovation Team
Huawei Turkey Research and Development Center, Istanbul

*Özetçe*

**Doğal dil işlemenin (NLP) temel görevlerinden biri olan bilgi çıkarma (IE), özgeçmişlerin kullanım alanındaki önemi son zamanlarda artış göstermiştir. Özgeçmişten bilgi çıkarmak için metin üzerinde yapılan çalışmalarda, genellikle NLP modellerini kullanarak cümle sınıflandırılması yapılmıştır. Bu çalışmada, özgeçmişlerin Optik Karakter Tanıma (OKT) ve YOLOv8 modeli ile nesne tanıma gibi ön işlemden sonra metin grupların tümünü sınıflandırarak bilgi çıkarımı hedeflenmektedir. Metin veri kümesi, bilişim sektöründeki 5 farklı (eğitim, tecrübe, yetenek, kişisel ve dil) iş tanımı için toplanan 286 özgeçmişten oluşmaktadır. Nesne tanıma için oluşturulan veri kümesi ise açık kaynak internetten toplanan ve metin kümeleri olarak etiketlenen 1198 özgeçmişten oluşmaktadır. Model olarak BERT, BERT-t, DistilBERT, RoBERTa ve XLNet kullanılmıştır. Model sonuçların karşılaştırılmasında F1 skor varyansları kullanılmıştır. Ayrıca YOLOv8 modeli de kendi içinde karşılaştırmalı olarak raporlanmıştır. Karşılaştırma sonucunda, DistilBERT diğer modellere göre daha az parametre sayısına sahip olmasına rağmen daha iyi sonuç göstermiştir.**

*Anahtar Kelimeler — Optik Karakter Tanıma (OKT), Doğal Dil İşleme, Bilgi çıkarma, Metin Sınıflandırma*

Abstract

**Information extraction (IE), one of the main tasks of natural language processing (NLP), has recently increased importance in the use of resumes. In studies on the text to extract information from the CV, sentence classification was generally made using NLP models. In this study, it is aimed to extract information by classifying all of the text groups after pre-processing such as Optical Character Recognition (OCT) and object recognition with the YOLOv8 model of the resumes. The text dataset consists of 286 resumes collected for 5 different (education, experience, talent, personal and language) job descriptions in the IT industry. The dataset created for object recognition consists of 1198 resumes, which were collected from the open-source internet and labeled as sets of text. BERT, BERT-t, DistilBERT, RoBERTa and XLNet were used as models. F1 score variances were used to compare the model results. In addition, the YOLOv8 model has also been reported comparatively in itself. As a result of the comparison, DistilBERT was showed better results despite having a lower number of parameters than other models.**

## I. GİRİŞ

Bilgi çıkarma (IE), doğal dil işleme (NLP) alanında önemli bir görev olarak kabul edilir. Özellikle, özgeçmişlerin bilgi çıkarma süreci, IE uygulamalarının en kritik senaryolarından birini oluşturur. Her bir özgeçmiş bölümünü sınıflandırarak metin verileri elde edilir, bu da sonraki arama ve analizler için kolay bir yöntemdir. Ayrıca, oluşturulan özgeçmiş verileri yapay zeka tabanlı özgeçmiş tarama sistemlerinde kullanılabilir, bu da insan kaynakları (HR) departmanları için işgücü maliyetini önemli ölçüde azaltabilir. Bu çalışma, özgeçmiş bilgi çıkarma görevini basit bir metin sınıflandırma görevine dönüştürme amacını taşımaktadır.

Literatürde özgeçmişlerden bilgi çıkarma çok farklı bakış açılarına sahip çalışmalar bulundurmaktadır [9,10,11,12]. Bu çalışmalardan birinde [10], yazarlar RoBERTa ve BERT gibi doğal dil işleme modelleri kullanarak cümle sınıflandırması yapmışlardır. Başka bir çalışma olan [9]'da yazarlar, prompt öğrenimi yöntemini özgeçmiş bilgi çıkarma görevine uygulamaya ve mevcut yöntemi özgeçmiş bilgi çıkarma görevine daha uygulanabilir hale getirmeye çalışmışlardır. Ayrıca, Maskeli Dil Modeli (MLM) ön eğitimli dil modelleri (PLM) ve Seq2Seq PLM'lerin bu görevdeki performansını karşılaştırmışlardır. Yazarlar, bilgili Prompt Ayarlama yöntemi için sözleyici tasarım yöntemini geliştirmiş ve diğer uygulama tabanlı NLP görevlerine yönelik Prompt şablonları ve sözleyici tasarımına bir örnek sağlamak amacıyla iyileştirmişlerdir. Bu bağlamda, Manuel Bilgili Sözleyici (MKV) kavramını öne sürmüşlerdir. Daha farklı olarak bilgi çıkarma alanında araştırma için [11]'de yazarlar, Çince büyük veri kümesi paylaşmış ve BERT, CNN gibi modeller ile veri kümesini eğitmişlerdir.

Optik Karakter Tanıma (OKT), bir metin görüntüsünü makine tarafından okunabilir bir metin biçimine dönüştüren işlemdir. OKT'nin temel amacı ise karakterleri maksimum doğrulukla tanımaktır. [19]'daki çalışmada yazarlar, OKT ve NLP kullanarak dosya sistemlerindeki belgelerin sorunsuz aranmasını sağlamayı amaçlamışlardır. Ayrıca, görüntü yakalama, görüntü geliştirme, görüntü tanımlama ve veri çıkarma gibi metotlar da kullanmışlardır. Başka bir çalışmada ise [20] yazarlar, makine öğrenimi (ML) algoritmaları ile yapılandırılmamış verilerin analiz edilmesi ve yorumlanmasını araştırmışlardır. Yazarlar, aynı zamanda yazma stili, kelime seçimi ve yapılandırılmamış sözdizimsel yazı dili ile ilgili araştırma kısıtlamalarını ve özgeçmiş analizinin gelecekteki potansiyelini tartışmışlardır.

Bu çalışmada özgeçmişten bilgi çıkarımı için birden fazla yaklaşım aynı potada eritilerek karşılaştırmalı bir yöntem sunulmuştur. Bu çalışma, varlık isim tanımlama, cümle sınıflandırması gibi diğer yaklaşımların aksine özgeçmişlerin Optik Karakter Tanıma ya da Metin Madenciliği ve nesne tanıma gibi birçok ön işlemden geçmesi sonrası metin grupların tümünü sınıflandırarak bilgi çıkarımını hedefler. Bu çalışmada sunulan yöntemin son adımı olan metin sınıflandırılması ele alınmıştır. Bu çalışmanın literatüre katkıları aşağıdaki gibi sıralanabilir.

(1) Özgeçmişte görüntüden metin gruplarının tespiti için açık kaynaklı verilerden etiketlenmiş bir açık kaynak veri kümesi sunulmuştur.
(2) Oluşturulmuş görüntü kümesi üzerinde YOLOv8 modeli eğitilmiş ve analizler gerçekleştirilmiştir.
(3) Açık kaynak olmayan metin veri kümesinde birbirinden farklı dil modelleri eğitilerek karşılaştırma yapılmış ve modellerin başarısı değerlendirilmiştir.

## II. VERİ

Veri kümesi bilişim sektöründeki 5 farklı iş tanımı için toplanmış 286 tane farklı anonimleştirilmiş ve genelleştirilmiş özgeçmişten oluşmaktadır. Veri kümesi eğitim, tecrübe, yetenek, kişisel ve dil olmak üzere 5 farklı bölüme ayrılmak maksadıyla bir insan tarafından etiketlenmiştir. Tüm veri etiketlenme boyunca özel isim ya da şirket ismi gibi özel kelimeler eşsiz tanımlarla yer değiştirilerek anonimleştirilmiştir. Veri kümesindeki tüm özgeçmişler dili İngilizce olarak belirlenmiştir. Veri toplama kısmında başka bir dil ile yazılmış özgeçmişler yok sayılırken, özgeçmişlerin içinde az sayıda geçen başka dile mensup kelimeler İngilizceye çevrilmiştir. Toplam 286 tane farklı özgeçmişten 1386 tane farklı bölüm vardır. Bu bölümlerden 286 tanesi kişisel, 282 tanesi eğitim, 281 tanesi yetenek, 272 tanesi tecrübe ve 231 tanesi dil olmak üzere 5 farklı sınıftan oluşmaktadır.

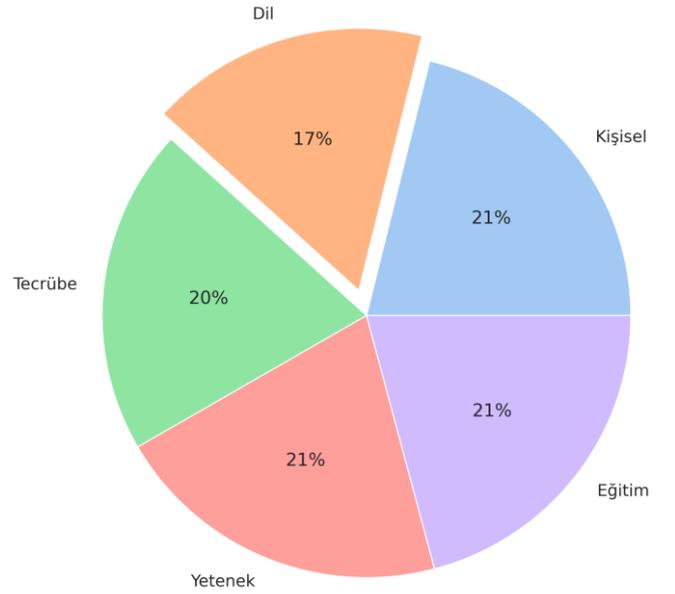

Şekil 1. Metin Veri Kümesindeki Bölüm Sayılarının Pasta Dağılımı

### A. Ön İşlem ve Veri Artırımı

Özgeçmişlerde büyük küçük ayrımını dikkate almamak için tüm kelimeler küçük harfe geçirildi. Özgeçmişlerde bölümlerin sınıf isimleri derin öğrenme modellerinde ezberleme gibi bir önyargı oluşturmasını önlemek adına veriden çıkarılmıştır. Veri kümesinde tutarlılık sağlamak adına sık kullanılan bağlaçlar ve noktalama işaretleri yok edilmiştir.

TABLO I. Tokenizasyon Şeması

| Model | Tokenizasyon | | | |
|---|---|---|---|---|
| | *Yöntem* | *Kelime Boyutu* | *Dolgulama* | *Maksimum Kelime Sayısı* |
| BERT | Word Piece | 30522 | Var | 75 |
| DistilBERT | Word Piece | 30522 | Var | 75 |
| RoBERTa | A byte-level BPE | 50265 | Var | 75 |
| XLNet | Sentence Piece | 32000 | Var | 75 |

Ayrıca, Tablo I. de görüldüğü gibi her omurga model yapısına uygun olacak şekilde tokenizasyon işlemi uygulanmış, maksimum kelime uzunluğu 75'e sabitlenmiştir. BERT ve DistilBERT modelleri için WordPiece [7] tokenizasyon yöntemi kullanırken, RoBERTa için Byte-Pair Encoding (BPE) ve XLNet için de Sentece Piece [2] tokenizasyon yöntemi kullanılmıştır.

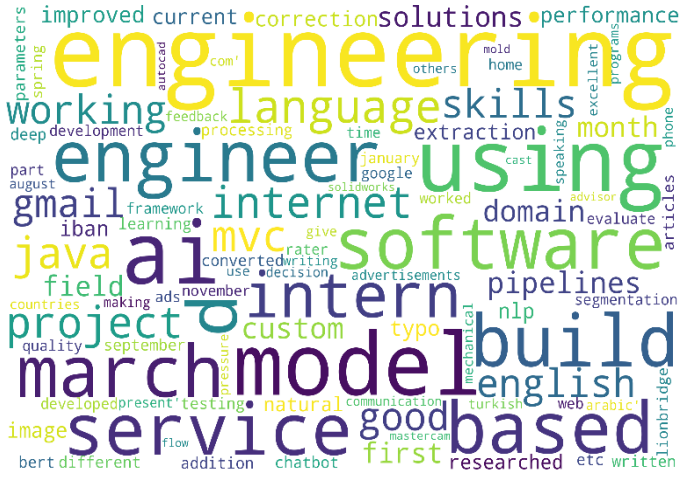

Şekil 2. Önişlemden Geçmiş Metin Veri Kümesinin Kelime Bulutu

Modellerin performanslarını artırmak, daha güçlü bir genelleme yapabilmesini sağlamak veya optik karakter tanıma yönteminden elde edilen verilerdeki hata payından kaçınmak gibi nedenlerden veri artırımı uygulanmış, veri sayısı 3 katını çıkarılmıştır.

Modellerin performanslarını artırmak, daha güçlü bir genelleme yapabilmesini sağlamak veya optik karakter tanıma yönteminden elde edilen verilerdeki hata payından kaçınmak gibi nedenlerden dolayı veri artırımı uygulanmış ve veri sayısı 3 katına çıkarılmıştır.

Veri artırmada rastgele olacak şekilde kelimelerdeki karakterlerin silinmesi, karakterlerin başka karakterler ile değiştirilmesi ve veriye yeni rastgele kelime eklenmesi yöntemleri kullanılmıştır. Bu yöntemlerin dışında verideki cümlelerin bağlamsal yapılarındaki doğallığını ve tutarlılığını koruyan iki farklı yöntem kullanılmıştır. Bunlardan ilki metin verisinde en iyi 100 benzer kelimeyi önceden eğitilmiş BERT modeli kullanarak bulur ve rastgele değiştirir. İkinci ise geri çeviri temelli işlem yaparak kelimeyi değiştirir. Örneğin, kaynak Türkçedir. Bu arttırıcı yöntemi İngilizce 'ye çevirir ve daha sonra geri Türkçe 'ye çevirir. Bu çevirilerde daha önce eğitilmiş modeller kullanıldı [1]. Veri artırımı boyunca tüm yöntemler rastgele dağılacak bir şekilde kullanılmıştır.

### B. Nesne Tanıma için Veri Kümesi

Veri kümesi[1] açık kaynak internetten toplanarak metin kümeleri olarak etiketlenmiş 1198 özgeçmişten oluşmaktadır. Bu özgeçmiş veri kümesi, diğer metin veri kümesinden bağımsız olarak toplanmıştır. Veri kümesinde her bir metin grubu nesne olarak kutucuk şekilde işaretlenmiştir. Ayrıca özgeçmişteki isim başlıkları yok sayılmıştır. Son olarak görüntülerin büyüklüğü 640x640 olacak şekilde sabitlenmiştir.

---
[1]https://app.roboflow.com/hwrd/resumes-4xmu5/6

[2]https://github.com/JaidedAI/EasyOCR

[3]PDF, platformlar arası taşınabilir ve yazdırılabilir belgeler oluşturmak amacıyla üretilmiş sayısal bir dosya biçimidir

[4]Microsoft'un tescilli Microsoft Word İkili Dosya Biçiminde depolanan sözcük işlem belgeleri için kullanılan bir dosya adı uzantısıdır

## III. YÖNTEM

Sunulan yöntem üç anahtar aşamadan oluşmaktadır. Birinci aşama özgeçmişten görüntü alınarak metin grup lokasyonunun belirlenmesi, ikinci aşama lokasyon bilgisinden görüntülerin kırpılarak önceden eğitilmiş OKT modeline gönderilerek metin karakterlerinin elde edilmesi, son aşama ise elde edilmiş metinin dil modelleriyle sınıflandırılmasıdır. Model diyagramı Şekil 3'de görülmektedir.

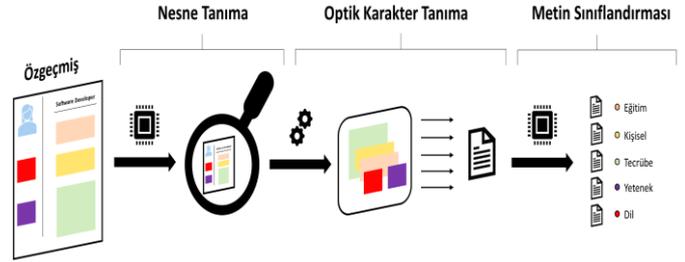

Şekil 3. Sunulan Yöntemin Diyagramı

### A. Optik Karakter Tanıma ve Nesne Tanıma

OKT (Optik Karakter Tanıma) yönteminden önce özgeçmişteki görüntü ele alınarak metin grupları etiketlenmelidir. Buradaki metin gruplarının ayrı ayrı etiketlenmesinde nesne tanıma yaklaşımı benimsenmiştir. Nesnelerin tanımı için YOLOv8 modeli eğitilmiştir [14].

Daha sonra eğitilmiş YOLOv8 model çıktıları OKT için önceden eğitilmiş modele verilerek metinler tanımlanır. OKT modeli Evrişimli Tekrarlayan Sinir Ağı mimari yapısına sahip önceden eğitilmiş bir modeldir [12,13]. Bu model, 3 temel bileşenden oluşmaktadır: özellik çıkarma (Resnet) [17], dizi etiketleme (LSTM) [16] ve çözümleme (CTC) işlemleri [15]. Önceden eğitim parametreleri EasyOCR[2] aracı kullanılarak elde edilmiştir. Fakat OKT dışında özgeçmişler PDF[3] ve DOC[4] uzantılı dosyalarda dosya kırpıldıktan sonra herhangi bir yapıyla bağdaştırmadan veya sınıflandırmadan yazı madenlenerek elde edilebilir [18].

### B. BERT

BERT (Bidirectional Encoder Representations from Transformers) [3], farklı iki yönlü dönüştürücü tabanlı bir derin öğrenme dil modelidir ve iki aşamada eğitilir. İlk aşamada etiketlenmemiş veriler ve maskelenmiş dil modellemesi kullanılarak sonraki cümleyi tahmin etme görevleri ile eğitilir. Daha sonra farklı dil problemleri için ek çıktı katmanları eklenerek modelin eğitimi sağlanır. BERT, doğal dil işleme alanında birçok görevde kullanılmaktadır, özellikle sözdizimi, anlambilim, duygu analizi, varlık ismi tanıma ve metin sınıflandırma gibi problemlerde yüksek performans sergilediği gözlenmiştir.

Bu çalışmada BERT modeli karşılaştırma amaçlı iki farklı varyasyonda kullanılmıştır. Birincisi önceden eğitilmiş model üzerine ince ayar (fine-tune) yapılırken, diğeri ise önceden eğitilmemiş model üzerine eğitilmiştir.

## C. DistilBERT

DistilBERT [4], BERT modelinin daha küçük ve hızlı çalışan bir versiyonudur. BERT modeline göre %40 daha az parametreye sahiptir ve %60 daha hızlı çalışabilir. Pooler ve token-tipi gömme katmanları çıkarılarak ve katman sayısı azaltılarak modelin boyutu küçültülmüştür. Böylece, BERT modeline yakın performansa sahip olmasına rağmen daha az parametre ve daha hızlı çalışma özellikleri taşıyan bir dil modeli elde edilmiş olur.

## D. RoBERTa

Robustly Optimized BERT Approach" (RoBERTa) [5], BERT modeline dayalı bir dil modelidir. RoBERTa, BERT modeline benzer bir çift yönlü dönüştürücü tabanlı dil modelidir ve çeşitli performans artırıcı optimizasyonlar ile eğitilmiştir. Daha geniş bir veri kümesi, uzatılmış eğitim süreleri ve daha büyük mini toplu boyutları kullanarak RoBERTa, BERT modeline kıyasla üstün sonuçlar elde eder. Ayrıca, RoBERTa, maskelenmiş dil modellemesi görevini daha hassas bir şekilde uygular ve cümlenin sırasını tahmin etme görevini de içerir.

## E. XLNet

XLNet, "Extra-Long Transformer" (Transformer) [6] mimarisine dayanan bir dil paradigmasıdır. Geleneksel otoregresif modellere karşın, XLNet, tüm cümleyi tahmin etmek üzere önceki kelimeler yerine cümle içindeki tüm kelime dizisini dikkate alarak çalışan bir otoregresif dil modelidir. Bu, modelin cümle içindeki ilişkileri daha iyi anlamasına ve daha doğru tahminler yapmasına olanak tanır. Sonuç olarak, XLNet, dil modelleme görevlerinde daha yüksek bir performans elde edebildiği ve daha uzun cümleler için daha doğru tahminler yapabildiği gözlenmiştir [6]. Ayrıca, XLNet, maskelenmiş dil modellemesi ve cümle sırası tahmini gibi ek görevleri de içerebilir, bu da geniş bir doğal dil işleme (NLP) görev yelpazesine uygulanabilir.

Sunulmuş son doğal dil işleme model yapısı iki anahtar modülden oluşmaktadır. Birinci omurga kısmı olarak BERT, DistilBERT, XLNet ve RoBERTa olarak 4 farklı versiyondan oluşurken, son modül ise iki lineer katmandan oluşan sınıflandırıcı başlığıdır. Sınıflandırıcı başlığında her lineer katman öncesi seyreltme katmanı (Dropout) bulunur.

## IV. DENEYLER

Veri seti önceden eğitilmiş modeller ile ince ayar edilmiştir. Omurga kısımları dondurularak yalnızca sınıflandırma başı eğitilmiştir. Yalnızca BERT modeli ayrıca karşılaştırma için tam eğitilmiştir.

Deneylerde tüm doğal dil işleme modeller yakınsama sınırına kadar eğitilmiştir. Modellerde iyileştirici (optimizer) olarak Adam kullanılırken öğrenme oranı 0.001'dir. Kayıp fonksiyonu ağırlıklı çapraz entropi olarak belirlenmiştir. Ağırlıklar sınıf sayılarının birbirine orantılı olarak hesaplanmıştır.

Sonuçların karşılaştırılmasında F1 skor varyansları kullanılmıştır. Bunlar sırasıyla F1 Mikro, F1 Makro ve F1 Ağırlıklı olmak üzere üç farklı versiyondur. Ayrıca sonuçların daha iyi gözlemlenebilmesi için hata matrisleri çıkarılmıştır. Veri metinleri kişi bazlı olarak eğitim, doğrulama ve test olarak üç kümeye ayrılmıştır. Test ve doğrulama metin veri kümesinde aynı kişiye ait olan metin eğitim kümesinde bulunmamaktadır. Kümelerin oranları sırasıyla eğitim %70, doğrulama %15, ve test %15 olacak şekilde ayrılmıştır.

TABLO II. F1 Skor Sonuçları

| Eğitim Planı | Model | F1 Skorlar | | |
|---|---|---|---|---|
| | | *F1 Mikro* | *F1 Makro* | *F1 Ağırlıklı* |
| İnce Ayar | BERT | 0.9391 | 0.9391 | 0.9388 |
| Tam Eğitim | BERT-t [a] | 0.9383 | 0.9391 | 0.9384 |
| İnce Ayar | **DistilBERT** | **0.9741** | **0.9746** | **0.9748** |
| İnce Ayar | RoBERTa | 0.9287 | 0.9289 | 0.9292 |
| İnce Ayar | XLNet | 0.9235 | 0.9239 | 0.9236 |

[a]. *İnce Ayarsız Tam Öğrenme ile Eğitilmiş BERT modeli*

Tablo 2' de görüldüğü gibi deneylerde en iyi performansı DistilBERT göstermiştir. Diğer modeller de DistilBERT'e yakın sonuçlar göstermiştir.

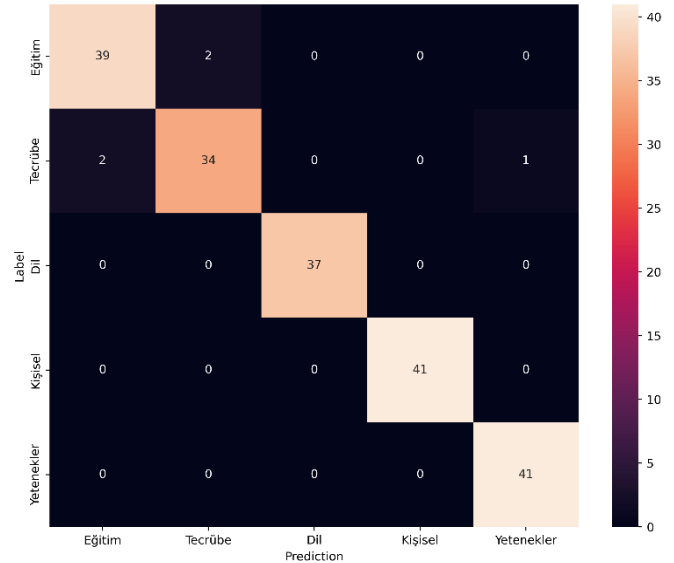

Şekil 3. DistilBERT Sonuçlarının Hata Matrisi

Nesne tanıma veri kümesi eğitim ve test olacak şekilde ikiye ayrılmıştır. Kümelerin oranları sırasıyla eğitim %85 test, %15 olacak şekilde ayrılmıştır. Yolov8 modeli yakınsama sınırına kadar 3 farklı versiyonda eğitilmiştir. Bu versiyonlar parametre sayısına göre değişiklik göstermektedir. Modelde iyileştirici olarak SGD kullanılırken öğrenme oranı 0.001'dir. Sonuçların karşılaştırılmasında ortalama hassasiyet (mAP - Mean Average Precision) kullanılmıştır.

TABLO III.    Nesne Tanıma Modelinin Sonuçları

| Model | Sonuçlar | |
|---|---|---|
| | $mAP^{50}$ | $mAP^{50-95}$ |
| Yolov8 Büyük | **0.8456** | 0.6362 |
| Yolov8 Orta | 0.8370 | **0.6390** |
| Yolov8 Küçük | 0.7970 | 0.5800 |

*A. Örnek Veri*

Örnek yaratılmış veri üzerinden tüm sunulan yöntem çalıştırılmıştır. Yaratılmış veri, metin ve nesne tanıma verisine yakınsanacak şekilde üretilmiştir. Buradaki amaç tüm yöntemin canlandırılmasıdır.

(1) Nesne Tanıma: PDF uzantılı özgeçmiş görüntüye çevrilmiştir. Görüntüye çevrildikten sonra nesne tanıma modeline verilmiştir. Görüntüde tespit edilen metin grupları kırpıldıktan sonra OKT uygulanmış metin yazıları elde edilmiştir.

(2) Elde edilen metin yazıları doğal dil işleme modeline verilerek özgeçmişten bilgi çıkarımı yapılmıştır.

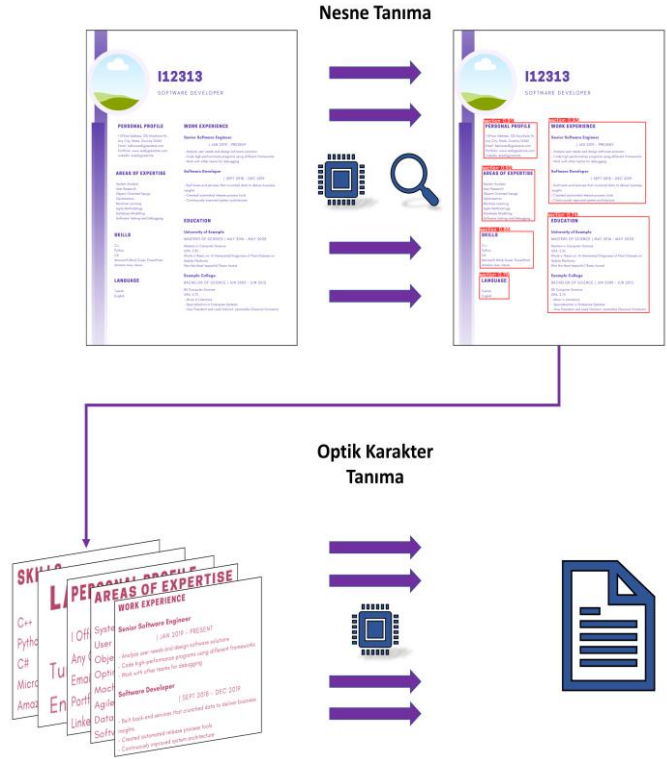

Şekil 4. Örnek Veri Üzerinde Önişlemler Diagramı

TABLO III.    Örnek Verinin Son Sonuçları

| Sınıf | Sonuç |
|---|---|
| Eğitim | EDUCATION<br>University of Example MASTERS OF SCIENCE MAY 2016 MAY 2020<br>Masters in Computer Science GPA: 3.35<br>Wrote a Thesis on: AI-Generated Diagnoses of Plant Disease for Mobile Platforms Won the Most Impactful Thesis Award<br>Example College BACHELOR OF SCIENCE JUN 2009 JUN 2012<br>BS Computer Science GPA: 3.75<br>Minor in Literature Specialization for Enterprise Systems |
| Tecrübe | WORK EXPERIENCE<br>Senior Software Engineer<br>JAN 2019 PRESENT<br>Analyze user needs and design software solutions Code high-performance programs using different frameworks Work with other teams for debugging<br>Software Developer<br>SEPT 2018 DEC 2019<br>Built back-end services that crunched data to deliver business insights Created automated release process tools Continuously improved system architecture |
| Yetenek | SKILLS<br>C++ Python C# Microsoft Word, Excel, PowerPoint Amazon Aws, Azure |
| Dil | LANGUAGE<br>Turkish English |
| Kişisel | PERSONAL PROFILE<br>Office Address: 123 Anywhere St. Any City, State, Country 12345 Email: ...@com Portfolio: www....com LinkedIn: example |

Şekil 4. ve Tablo 3. 'de görüldüğü gibi örnek veri üzerinde yöntemin simülasyon sonuçları paylaşılmıştır.

V. TARTIŞMA

Özgeçmişler yapılandırılmamış belgelerdir ve bu belgeleri büyük miktarlarda taramak zor bir işlemdir. Adayların özgeçmişlerini yapısal olarak oluşturmak, saklamak, işlemek ve gerektigindeki içeriklerindeki anlamlı bilgileri incelemek, işverenler için önemli bir zaman kazancı sağlar. Bu sayede işverenler özgeçmişleri, kapsamlı bir şekilde inceleyebilir ve ayıklayabilir.

Bu çalışmada, derin öğrenme teknikleriyle özgeçmişlerin bölümlerinin sınıflandırılarak tespit edilmesini ve böylece yapısal hale getirilmesine olanak sağlayan bir sistem sunulmaktadır. Yapılan kapsamlı deneylerde, DistilBERT diğer modellere göre daha az parametre sayısına sahip olmasına rağmen daha iyi sonuç vermiştir. Daha küçük bir model, daha az parametre ve daha az karmaşıklığa sahip olabilir, bu da modelin daha hızlı ve daha etkili öğrenmesini sağlayabilir. Aşırı öğrenme riski azalabilir ve genelleştirme yeteneği artabilir. Burada DistilBERT diğer büyük modellere göre küçük veri kümelerinde daha etkili bir eğitime sahip olduğu sonucu çıkarılabilir. Ayrıca önişlem aşamalarında YOLOv8 modeli iyi sonuçlar göstermiştir. Bunun yanında CRNN model [12]' de karşılaştırma çalışmasında umut verici sonuçlar göstermektedir. Sonuç olarak bu araştırmada, özgeçmiş özelinde ham veri işlenerek değerli bilgiler elde edilmiştir.

Gelecekte, sunulan yöntem için özgeçmiş veri kümesini çeşitlendirmeyi, farklı formatlar icerecek sekilde zenginleştirmeyi, işlenmemiş özgeçmişleri yapısal hale getirerek önerilen yöntemi otomize etmeyi ve çevrimiçi kullanmayı hedefliyoruz.

## KAYNAKLAR

[1] Ng, Nathan, et al. "Facebook FAIR's WMT19 news translation task submission." arXiv preprint arXiv:1907.06616 (2019).

[2] Taku Kudo and John Richardson. Sentencepiece: A simple and language independent subword tokenizer and detokenizer for neural text processing. arXiv preprint arXiv:1808.06226, 2018.

[3] Devlin, Jacob, et al. "Bert: Pre-training of deep bidirectional transformers for language understanding." arXiv preprint arXiv:1810.04805 (2018).

[4] Sanh, Victor, et al. "DistilBERT, a distilled version of BERT: smaller, faster, cheaper and lighter." arXiv preprint arXiv:1910.01108 (2019).

[5] Liu, Yinhan, et al. "Roberta: A robustly optimized bert pretraining approach." arXiv preprint arXiv:1907.11692 (2019).

[6] Lample, Guillaume, and Alexis Conneau. "Cross-lingual language model pretraining." arXiv preprint arXiv:1901.07291 (2019).

[7] Wu, Yonghui, et al. "Google's neural machine translation system: Bridging the gap between human and machine translation." arXiv preprint arXiv:1609.08144 (2016).

[8] Devlin, Jacob, et al. "Bert: Pre-training of deep bidirectional transformers for language understanding." arXiv preprint arXiv:1810.04805 (2018).

[9] Gan, Chengguang, and Tatsunori Mori. "A Few-shot Approach to Resume Information Extraction via Prompts." arXiv preprint arXiv:2209.09450 (2022).

[10] Gan, Chengguang, and Tatsunori Mori. "Construction of English Resume Corpus and Test with Pre-trained Language Models." arXiv preprint arXiv:2208.03219 (2022).

[11] Y. Su, J. Zhang and J. Lu, "The Resume Corpus: A Large Dataset for Research in Information Extraction Systems," 2019 15th International Conference on Computational Intelligence and Security (CIS), Macao, China, 2019, pp. 375-378, doi: 10.1109/CIS.2019.00087.

[12] Baek, Jeonghun, et al. "What is wrong with scene text recognition model comparisons? dataset and model analysis." Proceedings of the IEEE/CVF international conference on computer vision. 2019.

[13] Shi, Baoguang, Xiang Bai, and Cong Yao. "An end-to-end trainable neural network for image-based sequence recognition and its application to scene text recognition." IEEE transactions on pattern analysis and machine intelligence 39.11 (2016): 2298-2304.

[14] Jocher, G., Chaurasia, A., & Qiu, J. (2023). YOLO by Ultralytics (Version 8.0.0) [Computer software]. https://github.com/ultralytics/ultralytics

[15] Graves, Alex, et al. "Connectionist temporal classification: labelling unsegmented sequence data with recurrent neural networks." Proceedings of the 23rd international conference on Machine learning. 2006.

[16] Hochreiter, Sepp, and Jürgen Schmidhuber. "Long short-term memory." Neural computation 9.8 (1997): 1735-1780.

[17] He, Kaiming, et al. "Deep residual learning for image recognition." Proceedings of the IEEE conference on computer vision and pattern recognition. 2016.

[18] Vidhya, K. A., & Aghila, G. (2010). Text mining process, techniques and tools: an overview. International Journal of Information Technology and Knowledge Management, 2(2), 613-622.

[19] A. M. Abbas, M. S. S. Hameed, S. Balakrishnan and K. S. Anandh, "Intelligent Document Finding using Optical Character Recognition and Tagging," 2022 International Conference on Automation, Computing and Renewable Systems (ICACRS), Pudukkottai, India, 2022, pp. 1165-1168, doi: 10.1109/ICACRS55517.2022.10029142.

[20] Pimpalkar, Amit, et al. "Job Applications Selection and Identification: Study of Resumes with Natural Language Processing and Machine Learning." 2023 IEEE International Students' Conference on Electrical, Electronics and Computer Science (SCEECS). IEEE, 2023.